\documentclass{article} 
\usepackage{iclr2027_conference,times}

\usepackage{amsmath,amsfonts,bm}

\def\eqref#1{equation~\ref{#1}}

\def\1{\bm{1}}

\DeclareMathAlphabet{\mathsfit}{\encodingdefault}{\sfdefault}{m}{sl}
\SetMathAlphabet{\mathsfit}{bold}{\encodingdefault}{\sfdefault}{bx}{n}

\usepackage{hyperref}
\usepackage{url}
\usepackage{algorithm}
\usepackage{algpseudocode}
\usepackage{booktabs}
\usepackage{graphicx}
\usepackage{amsmath}
\usepackage{amssymb}
\usepackage{booktabs}
\usepackage{multirow}

\title{SkipWorld: Latent Skip Planning with Distilled World Models for Efficient Visual Navigation}

\author{
\textbf{Yifei Yang}$^{1,3,4}$ \quad
\textbf{Zehua Fan}$^{2}$ \quad
\textbf{Zike Yan}$^{3,4}$ \quad
\textbf{Zhongyu Li}$^{3,4}$ \quad
\textbf{Yan Wang}$^{1}$\thanks{Corresponding author.} \\
$^{1}$Institute for AI Industry Research (AIR), Tsinghua University \quad
$^{2}$Shanghai Jiao Tong University \\
$^{3}$Hong Kong Embodied AI Lab \quad
$^{4}$The Chinese University of Hong Kong
}

\iclrfinalcopy 
\begin{document}

\maketitle

\begin{abstract}

World-model-based visual navigation enables action planning by predicting future states under candidate trajectories. However, long-horizon rollout remains challenging: models are often trained with ground-truth visual histories but deployed in closed-loop prediction, and evaluating many candidates with step-by-step rollout is computationally expensive. We propose \textbf{SkipWorld}, a latent skip planning framework for efficient visual navigation. SkipWorld introduces three components: (i) \textbf{Latent AutoRoll}, which trains a world model to recursively roll out future states in latent space, reducing the mismatch between training and closed-loop inference; (ii) \textbf{SkipDistill}, which distills the latent autoregressive model into a skip world model that directly predicts long-stride future latent states; and (iii) \textbf{Skip-then-Refine Planning}, a dual-model coarse-to-fine planner that first filters candidate trajectories with the skip model and then reranks selected candidates with the stronger autoregressive model. This design preserves accurate trajectory evaluation while reducing redundant rollout computation. Experiments on RECON, HuRoN, SCAND, and TartanDrive show that, compared with NWM, SkipWorld reduces the average ATE by 4.2\% and RPE by 2.7\%, while decreasing planning time by 62.4\% with a 2.7$\times$ speedup.
\end{abstract}

\section{Introduction}

Visual navigation requires an agent to anticipate how its actions may change future observations and whether those changes will lead it closer to the goal. Rather than directly mapping observations to actions, world-model-based methods learn to predict future states and use these predictions for planning~\cite{bar2025navigation,huang2026unemo}. Given the current observation, goal, and candidate action trajectories, the model evaluates their possible outcomes and selects the sequence that best supports the navigation objective~\cite{bar2025navigation,chahe2026policy,shah2023vint}. This explicit look-ahead is especially useful for long-horizon navigation, where reactive policies may struggle with delayed consequences~\cite{shah2023vint,garg2025objectreact,shah2023gnm,sridhar2024nomad}.

Despite this promise, efficient planning with visual world models remains difficult. A central challenge is the mismatch between training and deployment. During training, many predictive models condition on ground-truth observation histories, where each future prediction receives clean visual context from the dataset. During planning, however, the model must operate in a closed loop: once the first future state is predicted, subsequent predictions depend on previously predicted states rather than ground-truth observations~\cite{bar2025navigation}. This discrepancy can cause errors to accumulate over long horizons, degrading the reliability of predicted trajectories and weakening downstream action selection~\cite{huang2026self}.

\begin{figure}[h]
    \centering
    \includegraphics[width=0.98\linewidth]{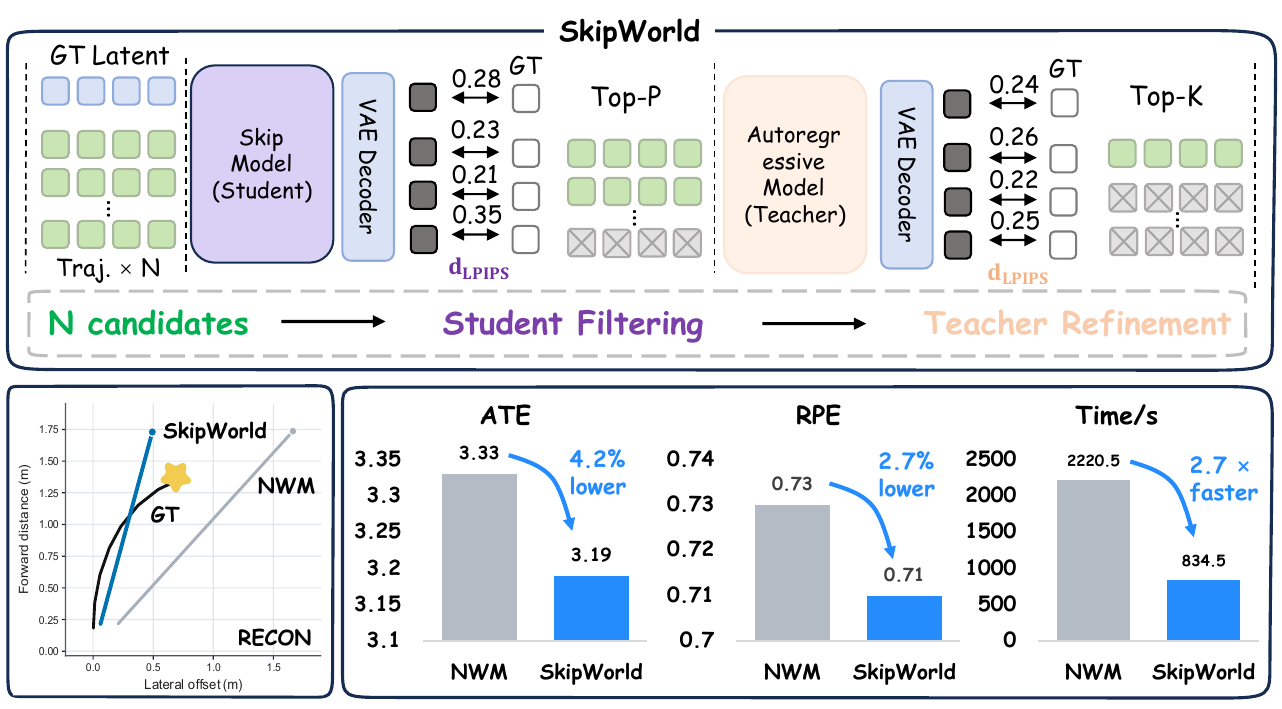}
    \caption{\textbf{Overview and performance of SkipWorld.} \textbf{Top:} SkipWorld performs coarse-to-fine planning by first using a skip student to filter $N$ candidate trajectories into a compact Top-$P$ subset, followed by latent autoregressive teacher refinement. \textbf{Bottom-left:} Example trajectory prediction on RECON, where SkipWorld more closely follows the ground-truth trajectory than NWM. \textbf{Bottom-right:} Compared with NWM, SkipWorld reduces average ATE by 4.2\% and RPE by 2.7\%, while achieving a $2.7\times$ planning speedup.}
    \label{fig:candidate_filtering}
\end{figure}

A second challenge is computational cost. Planning requires evaluating many candidate action trajectories, and each candidate may require multiple autoregressive prediction steps. When the model predicts or decodes high-dimensional images at every step, the cost grows quickly with the planning horizon, the number of samples, and the number of repeated evaluations. As a result, accurate world-model planning can become too slow for practical visual navigation, where decisions must be made repeatedly and efficiently~\cite{hansen2024td,chahe2026policy}.

One natural way to reduce this cost is to move prediction from pixel space to latent space. Latent representations provide a compact state abstraction that can preserve task-relevant visual and geometric information while avoiding unnecessary image reconstruction during planning~\cite{zhou2024dino,chahe2026policy}. However, simply encoding observations into a latent space is not sufficient. If the model is still trained with clean ground-truth histories but evaluated using its own predictions, the closed-loop mismatch remains. Likewise, if every candidate is still rolled out one step at a time, planning can remain expensive even when each step is cheaper.

To address these issues, we propose \textbf{SkipWorld} (Figure~\ref{fig:candidate_filtering}), a latent skip planning framework for efficient visual navigation. SkipWorld consists of three components. First, \textbf{Latent AutoRoll} trains a world model with closed-loop latent rollout, reducing the mismatch between training and inference~\cite{hafner2025mastering,zhou2024dino,yu2025unisim}. Second, \textbf{SkipDistill} distills this autoregressive model into a skip world model that directly predicts long-stride future latent states, enabling efficient trajectory evaluation. Third, \textbf{Skip-then-Refine Planning} combines the two models in a coarse-to-fine planner: the skip model rapidly filters a large candidate set, while the autoregressive model refines only promising trajectories~\cite{schrittwieser2021online,hansen2024td}. Together, these components improve the efficiency-accuracy trade-off of long-horizon visual world-model planning.

Our contributions are summarized as follows:
\begin{itemize}
\item We introduce \textbf{SkipWorld}, a unified latent planning framework that integrates closed-loop latent prediction, skip distillation, and coarse-to-fine candidate evaluation.

\item We develop \textbf{Latent AutoRoll} and \textbf{SkipDistill} to model long-horizon dynamics efficiently, combining autoregressive latent rollout with direct long-stride prediction.

\item We propose \textbf{Skip-then-Refine Planning}, which allocates expensive autoregressive rollout only to promising candidate trajectories.

\item \textbf{SkipWorld} reduces average ATE by 4.2\%, RPE by 2.7\%, and planning time by 62.4\% compared with NWM.
\end{itemize}

\section{Related Work}

\paragraph{Visual navigation.}
Learning-based visual navigation aims to infer goal-directed behaviors directly from egocentric observations. Large-scale training on heterogeneous robot trajectories has substantially improved cross-environment and cross-embodiment generalization~\cite{shah2023gnm,shah2023vint}, while diffusion-based policies further enable multimodal navigation behaviors for both goal-directed navigation and exploration~\cite{sridhar2024nomad}. Despite their strong generalization, such policy-based approaches typically map observations directly to actions without explicitly reasoning about the visual consequences of candidate trajectories~\cite{bar2025navigation,zhou2024dino}. This limitation has motivated predictive approaches that evaluate potential actions through their anticipated future observations~\cite{hafner2025mastering,hao2026latentpilot}.

\paragraph{World Models for Planning.}
World models enable decision making by predicting future states conditioned on candidate actions, and have been extensively studied for planning and control from visual observations~\cite{hafner2019learning,hansen2024td,hafner2025mastering}. In visual navigation, \textbf{Navigation World Models (NWM)}~\cite{bar2025navigation} formulate goal-conditioned navigation as visual foresight, predicting future observations under candidate action sequences and ranking them according to their proximity to the goal. Subsequent work has explored richer predictive representations, memory mechanisms, and alternative action parameterizations~\cite{dong2025unified,mei2026navwm,stapf2026composition}. \textbf{RAE-NWM}~\cite{zhang2026rae} performs navigation world modeling in a dense representation space, while \textbf{Learning Latent Action World Models in the Wild}~\cite{garrido2026learning} learns latent action dynamics from large-scale unlabeled video and transfers them to navigation through lightweight action alignment. \textbf{Beyond Language Modeling}~\cite{tong2026beyond} further studies how multimodal pretraining data affect world-model-based navigation planning. Despite these advances, candidate trajectories are still commonly evaluated through multi-step sequential prediction, making planning increasingly expensive as the number of candidates and rollout horizon grow.

\paragraph{Efficient Candidate Evaluation.}
Efficient model-based planning has traditionally reduced computation through compact latent dynamics, efficient trajectory optimization, or learned predictive representations~\cite{hafner2019learning,hansen2024td}. In navigation, \textbf{NWM}~\cite{bar2025navigation} evaluates a large population of candidate action sequences by repeatedly rolling the world model forward, which provides detailed visual foresight but incurs substantial redundant computation. Recent approaches improve efficiency by reducing the computational cost of the underlying representation or rollout model~\cite{zhang2026rae,stapf2026composition}. In particular, \textbf{CompACT}~\cite{kim2026planning} compresses observations into a small number of discrete tokens, substantially accelerating world-model inference while retaining step-by-step candidate rollout. These approaches primarily reduce the cost of individual prediction steps, whereas the number of sequential predictions required for evaluating each candidate largely remains unchanged.

\section{Method}

\begin{figure}[t]
    \centering
    \includegraphics[width=\linewidth]{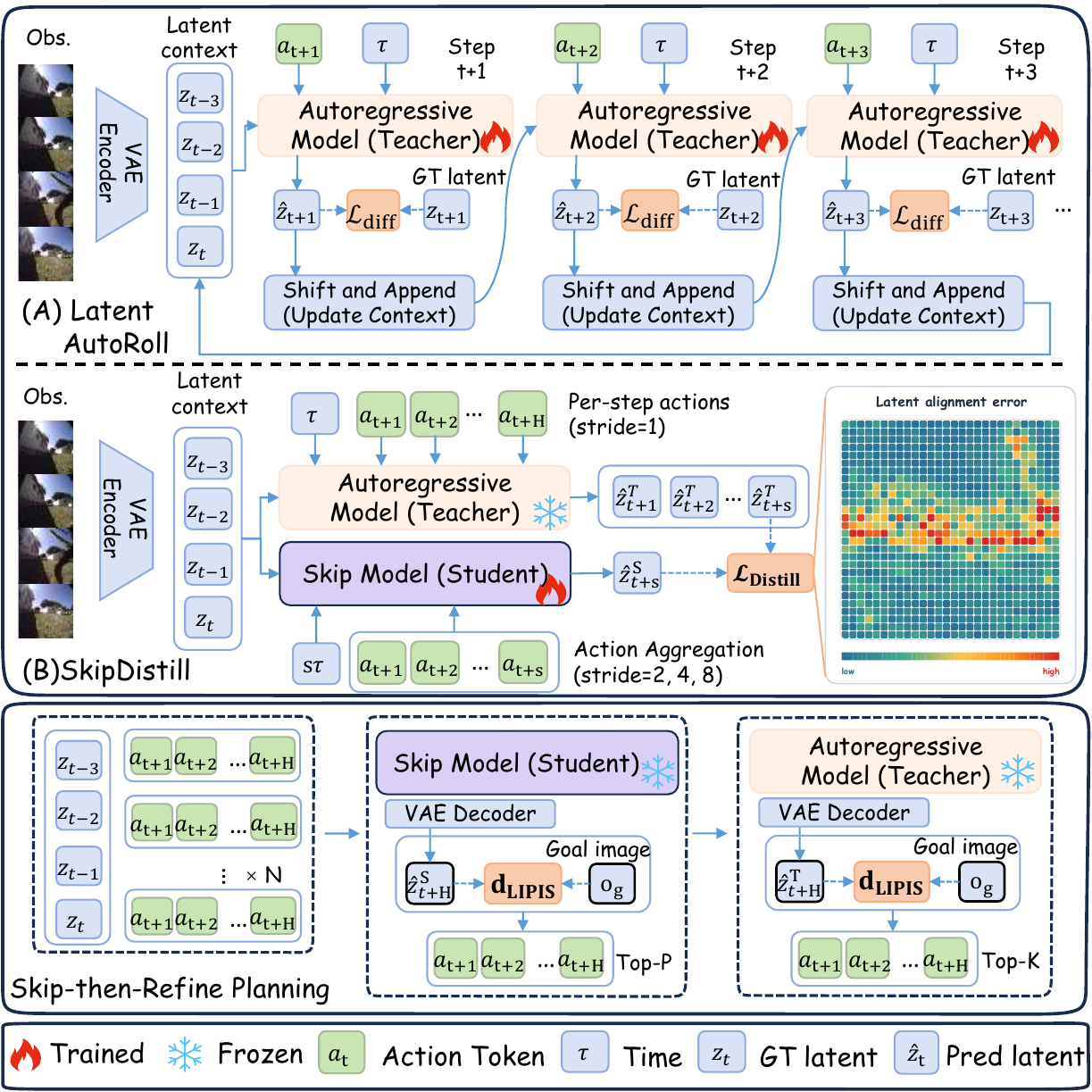}
    \caption{Overview of SkipWorld. First, \textbf{Latent AutoRoll} trains an autoregressive world model that rolls out directly in latent space. Second, \textbf{SkipDistill} trains a skip student to predict longer-horizon latent transitions through teacher endpoint distillation, where the latent alignment error (Section~\ref{app:latent_alignment_error}) visualizes the spatial discrepancy between the student skip endpoint and the autoregressive teacher endpoint. Third, \textbf{Skip-then-Refine Planning} uses the skip student to cheaply filter candidate trajectories and then uses the autoregressive teacher to accurately rerank only the selected candidates.}
    \label{fig:method}
\end{figure}

\subsection{Problem Formulation}

As illustrated in Fig.~\ref{fig:method}, SkipWorld performs
world-model-based visual navigation by predicting future observations
under candidate action trajectories and selecting actions according to
their visual discrepancy to the goal.

Let the robot observe a context window of RGB frames and a goal image.
We first encode each observation into latent tokens using a pretrained
VAE encoder~\cite{blattmann2023stable}, i.e., $z_t=E(o_t)$.
The latent context is represented as
$C_t=[z_{t-c+1},\ldots,z_t]$, where $c$ denotes the context size.
We use $t$ to denote the current time step and
$h\in\{1,\ldots,H\}$ to index future rollout steps.

The world model predicts the next latent state conditioned on the latent
context, the corresponding action, and the relative temporal interval,
i.e.,
$\hat{z}_{t+1}=F_\theta(C_t,a_{t+1},\tau)$,
where $F_\theta$ denotes the latent diffusion world model and $\tau$
is the relative temporal condition. In our implementation, the
denoising network is a conditional diffusion transformer~\cite{bar2025navigation}.

During planning, each candidate action trajectory
$a_{t+1:t+H}=[a_{t+1},\ldots,a_{t+H}]$ is evaluated by rolling out the
world model into the future. The final predicted latent
$\hat{z}_{t+H}$ is decoded into image space using the VAE decoder
$D(\cdot)$ and compared with the goal image $o_g$ using the perceptual
distance $d_{\mathrm{LPIPS}}$~\cite{zhang2018unreasonable}. This
distance is used as the planning cost for candidate ranking.

\subsection{Latent AutoRoll}

The first component, Latent AutoRoll, trains the world model under the
same closed-loop latent rollout pattern used during inference.
Standard world-model training typically conditions future predictions
on ground-truth histories, whereas planning recursively feeds the
model's own predictions back into the context. This discrepancy can
cause prediction errors to accumulate over long horizons.

Latent AutoRoll explicitly exposes the model to its own predicted
latents during training. Starting from the current latent context
$C_t$, the model recursively predicts future latent states as

\begin{equation}
\hat{z}^{T}_{t+h}
=
F_\theta
\left(
C_{t+h-1},
a_{t+h},
\tau
\right),
\qquad
h=1,\ldots,H.
\end{equation}

After each prediction, the context is updated as
$C_{t+h}=\mathrm{shift}(C_{t+h-1},\hat{z}^{T}_{t+h})$,
where $\mathrm{shift}(\cdot)$ removes the oldest latent from the context
window and appends the predicted clean latent as the newest context
state. Therefore, predictions at later rollout steps are conditioned
on previous model-generated latents rather than ground-truth future
states, matching the closed-loop rollout used during planning.

For a training sequence with future target latents
$\{z_{t+1},\ldots,z_{t+H}\}$, the Latent AutoRoll objective is

\begin{equation}
\mathcal{L}_{\mathrm{AR}}
=
\frac{1}{H}
\sum_{h=1}^{H}
\mathcal{L}_{\mathrm{diff}}
\left(
\theta;
z_{t+h},
C_{t+h-1},
a_{t+h},
\tau
\right),
\end{equation}

where $\mathcal{L}_{\mathrm{diff}}$ denotes the diffusion denoising
objective and $z_{t+h}$ is the ground-truth target latent at rollout
step $h$. By recursively training on predicted latent contexts,
Latent AutoRoll enables the world model to recover from its own
prediction errors and reduces the mismatch between training and
planning-time rollout.

\subsection{SkipDistill}

Although Latent AutoRoll improves consistency between training and
closed-loop inference, dense autoregressive rollout remains expensive.
If every candidate trajectory requires $H$ sequential one-step model
calls, planning cost grows rapidly with both the rollout horizon and
the number of sampled candidates.

SkipDistill addresses this issue by distilling the dense autoregressive
teacher into a skip student that directly predicts a future latent
endpoint over a larger temporal interval. For a sampled skip stride
$s\in\{2,4,8\}$, the actions over the skipped interval are accumulated
as $\bar{a}_{t,s}=\sum_{i=1}^{s}a_{t+i}$.
The student predicts the endpoint latent from the current context, the
accumulated action, and the corresponding temporal displacement:

\begin{equation}
\hat{z}^{S}_{t+s}
=
F_\phi
\left(
C_t,
\bar{a}_{t,s},
s\tau
\right),
\end{equation}

where $F_\phi$ denotes the skip student,
$\hat{z}^{S}_{t+s}$ is the student-predicted endpoint latent,
$\bar{a}_{t,s}$ represents the accumulated action over the skipped
interval, and $s\tau$ is the scaled relative temporal condition.

The student is supervised by the endpoint produced by the frozen
Latent AutoRoll teacher. Starting from the same context $C_t$, the
teacher performs $s$ dense one-step predictions,

\begin{equation}
\hat{z}^{T}_{t+j}
=
F_\theta
\left(
C^{T}_{t+j-1},
a_{t+j},
\tau
\right),
\qquad
j=1,\ldots,s,
\end{equation}

with the context recursively updated as
$C^{T}_{t+j}
=
\mathrm{shift}(C^{T}_{t+j-1},\hat{z}^{T}_{t+j})$.
This dense rollout produces the teacher endpoint
$\hat{z}^{T}_{t+s}$, which serves as the distillation target for the
student.

The endpoint distillation objective is

\begin{equation}
\mathcal{L}_{\mathrm{distill}}
=
\mathrm{SmoothL1}
\left(
\hat{z}^{S}_{t+s},
\mathrm{sg}
\left(
\hat{z}^{T}_{t+s}
\right)
\right),
\end{equation}

where $\mathrm{sg}(\cdot)$ denotes stop-gradient and
$\mathrm{SmoothL1}$~\cite{girshick2015fast} is the robust regression loss used in
implementation. By matching the endpoint of the dense teacher rollout,
the skip student learns to approximate multiple autoregressive steps
with a single prediction, substantially reducing sequential inference
cost during candidate screening.

\subsection{Skip-then-Refine Planning}

At inference time, SkipWorld combines the efficient skip student and
the stronger Latent AutoRoll teacher in a coarse-to-fine planning
strategy. The planner first samples $N$ candidate action trajectories
over a horizon of $H$ steps.

In the coarse stage, the skip student evaluates all candidates using
stride $s$. For each candidate $a^i$, the student recursively performs
skip predictions until the planning horizon is reached. The final
predicted latent $\hat{z}^{S,i}_{t+H}$ is decoded and compared with the
goal image using

\begin{equation}
J_{\mathrm{skip}}(a^i)
=
d_{\mathrm{LPIPS}}
\left(
D(\hat{z}^{S,i}_{t+H}),
o_g
\right).
\end{equation}

The $P$ candidates with the lowest coarse costs are retained for dense
refinement. This coarse filtering stage avoids applying the expensive
autoregressive teacher to candidates that are unlikely to reach the
goal.

The retained trajectories are then evaluated by the Latent AutoRoll
teacher using dense stride-$1$ rollout. Since diffusion-based world
models are stochastic, each retained candidate is evaluated with $R$
independent rollouts. The refined planning score is

\begin{equation}
J_{\mathrm{AR}}(a^i)
=
\frac{1}{R}
\sum_{r=1}^{R}
d_{\mathrm{LPIPS}}
\left(
D(\hat{z}^{T,i,r}_{t+H}),
o_g
\right),
\end{equation}

where $\hat{z}^{T,i,r}_{t+H}$ denotes the final latent predicted for
candidate $a^i$ in the $r$-th dense teacher rollout.

The retained trajectories are reranked according to
$J_{\mathrm{AR}}$. The top-$K$ candidates with the lowest refined
costs are selected, and their actions are averaged to obtain the final
action. In this way, the skip student performs efficient coarse
screening over the full candidate set, while the stronger
autoregressive teacher is reserved only for the most promising
trajectories.

\begin{algorithm}[t]
\caption{Skip-then-Refine Planning}
\label{alg:skipworld_planning}
\begin{algorithmic}[1]

\Require observation context $o_c$, goal image $o_g$,
skip student $F_\phi$, autoregressive teacher $F_\theta$,
number of candidates $N$, retained candidates $P$,
top candidates $K$, horizon $H$, skip stride $s$,
stochastic repeats $R$

\State Encode observation context into latent context $C_t$
\State Sample $N$ candidate action trajectories
$\mathcal{A}_N=\{a^1,\ldots,a^N\}$

\For{each candidate $a^i \in \mathcal{A}_N$}
    \State Accumulate actions over each $s$-step interval
    \State Roll out the skip student $F_\phi$ with stride $s$
    \State Decode the final predicted latent
    \State Compute $J_{\mathrm{skip}}(a^i)$ using LPIPS
\EndFor

\State Retain the best $P$ candidates $\mathcal{A}_P$
according to $J_{\mathrm{skip}}$

\For{each candidate $a^i \in \mathcal{A}_P$}
    \For{$r=1,\ldots,R$}
        \State Roll out the teacher $F_\theta$ with stride $1$
        \State Decode the final predicted latent
        \State Compute the LPIPS cost
    \EndFor
    \State Average the $R$ costs to obtain $J_{\mathrm{AR}}(a^i)$
\EndFor

\State Select the top-$K$ candidates according to $J_{\mathrm{AR}}$
\State \Return the mean action of the top-$K$ candidates

\end{algorithmic}
\end{algorithm}

\subsection{Complexity Analysis}

We compare planning complexity in terms of world-model rollout calls.
Let $N$ be the number of sampled candidate trajectories, $H$ the
prediction horizon, $s$ the skip stride, $P$ the number of candidates
retained after coarse screening, and $R$ the number of stochastic
repeats used during dense refinement.

The original dense planner evaluates every candidate using stride-$1$
autoregressive rollout. Since each of the $N$ candidates requires $H$
world-model calls for each of the $R$ stochastic evaluations, its
rollout complexity is

\begin{equation}
\mathcal{C}_{\mathrm{NWM}}
=
O(NRH).
\end{equation}

SkipWorld replaces the dense rollout used during coarse candidate
evaluation with the skip student. Evaluating all candidates with stride
$s$ requires approximately $H/s$ student calls per trajectory, while
the expensive dense teacher is used only for the retained $P$
candidates. The resulting complexity is

\begin{equation}
\mathcal{C}_{\mathrm{SkipWorld}}
=
O\left(
N\frac{H}{s}
\right)_{\mathrm{student}}
+
O(PRH)_{\mathrm{teacher}}.
\end{equation}

With $N=120$, $H=8$, $s=8$, $P=20$, and $R=3$, the dense planner
requires 2880 dense rollout steps, whereas SkipWorld uses 120
skip-student steps and 480 dense teacher steps. This reduces expensive
teacher rollout computation by $6\times$. In practice, the reduced
rollout cost translates to a $2.7\times$ end-to-end speedup, with the
remaining overhead mainly arising from latent decoding, LPIPS
scoring~\cite{zhang2018unreasonable}, data movement, and prediction
saving.

\begin{table*}[t]
\centering
\caption{Main comparison on four visual navigation benchmarks. Lower is better. The best result is shown in bold and the second best is underlined. NWM (original) is from its original paper. NWM (pretrained) denotes our four‑dataset retrained baseline. Due to privacy restrictions, only a low‑resolution HuRoN subset is available. For fair comparison, We thus fine‑tune it from the 100K‑step checkpoint for another 50K steps with identical settings.}
\label{tab:main_results}
\setlength{\tabcolsep}{5pt}
\small
\begin{tabular}{lcccccccccc}
\toprule
\multirow{2}{*}{Model}
& \multicolumn{2}{c}{RECON}
& \multicolumn{2}{c}{HuRoN}
& \multicolumn{2}{c}{TartanDrive}
& \multicolumn{2}{c}{SCAND}
& \multicolumn{2}{c}{Avg.} \\
\cmidrule(lr){2-3}
\cmidrule(lr){4-5}
\cmidrule(lr){6-7}
\cmidrule(lr){8-9}
\cmidrule(lr){10-11}
& ATE & RPE & ATE & RPE & ATE & RPE & ATE & RPE & ATE & RPE \\
\midrule
GNM
& 1.87 & 0.73
& \underline{3.71} & 1.00
& 6.65 & 1.62
& 2.12 & 0.61
& 3.59 & 0.99 \\
NoMaD
& 1.95 & 0.53
& 3.73 & 0.96
& 6.32 & 1.31
& 2.24 & 0.49
& 3.56 & 0.82 \\
NWM + NoMaD ($\times16$)
& 1.88 & 0.51
& 3.73 & \underline{0.95}
& 6.26 & 1.30
& 2.18 & 0.48
& 3.51 & 0.82 \\
NWM + NoMaD ($\times32$)
& 1.79 & 0.49
& 3.68 & \underline{0.95}
& 6.25 & 1.29
& 2.19 & 0.47
& 3.47 & 0.80 \\
NWM (original)
& \textbf{1.13} & \underline{0.35}
& 4.12 & 0.96
& \textbf{5.63} & \textbf{1.18}
& 1.28 & 0.33
& \textbf{3.04} & \textbf{0.71} \\
CompAct
& 1.33 & 0.39
&  - &  -
&  - &  -
& 1.36 & 0.34
& - &  - \\
RAE-NWM
& 1.36 & 0.37
& \textbf{2.91} & \textbf{0.70} 
& - & -
& 1.14 & \textbf{0.28}
& - &  - \\

\midrule

NWM (pretrained)
& 1.38 & \underline{0.35}
& 5.08 & 1.08
& 5.79 & 1.21
& \underline{1.08} & \underline{0.29}
& 3.33 & \underline{0.73} \\
\textbf{SkipWorld}
& \underline{1.19} & \textbf{0.34}
& 4.86 & 1.04
& \underline{5.74} & \underline{1.20}
& \textbf{0.98} & \textbf{0.28}
& \underline{3.19} & \textbf{0.71} \\
\bottomrule
\end{tabular}
\end{table*}

\section{Experiments}

\subsection{Experimental Setup}

\paragraph{Benchmark.}
We evaluate SkipWorld on four visual navigation benchmarks: RECON~\cite{shah2021rapid}, HuRoN~\cite{hirose2023sacson},TartanDrive~\cite{triest2022tartandrive}, and SCAND~\cite{karnan2022socially}.  These benchmarks cover diverse navigation scenarios, including indoor navigation, human-centered navigation, off-road driving, and social navigation. We follow the standard planning evaluation protocol used by prior navigation world model methods.

\paragraph{Evaluation Metrics.}
We report absolute trajectory error (ATE)~\cite{sturm2012evaluating} and relative pose error (RPE)~\cite{sturm2012evaluating}. ATE measures global trajectory deviation, while RPE measures local motion consistency. Lower values indicate better performance.

\paragraph{Baselines.}
We compare SkipWorld with representative visual navigation methods.
\textbf{GNM}~\cite{shah2023gnm} is a general goal-conditioned navigation model trained on heterogeneous robot data.
\textbf{NoMaD}~\cite{sridhar2024nomad} employs a unified diffusion policy for goal-directed navigation and exploration.
\textbf{NWM}~\cite{bar2025navigation} is a navigation world model that evaluates action candidates by autoregressively predicting future visual observations.
\textbf{CoMPACT}~\cite{kim2026planning} performs efficient world-model planning using compact discrete visual tokens.
\textbf{RAE-NWM}~\cite{zhang2026rae} models navigation dynamics in a dense visual representation space using a representation autoencoder, improving future prediction and downstream planning.

\paragraph{Implementation Details.}

For \textbf{Latent AutoRoll}, we train for 50K steps from the 100K-step NWM checkpoint with a 16-frame horizon on four datasets. For \textbf{SkipDistill}, the student starts from the same 100K-step checkpoint and is distilled from the frozen 150K-step teacher using $\mathcal{S}={2,4,8}$ for 50K steps. For \textbf{planning}, unless otherwise stated, the planner samples $N=120$ candidates, keeps $K=5$ elites, and performs one CEM update step. The default SkipWorld configuration uses coarse stride $s_c=8$, preselection size $P=20$, and $R=3$ teacher repeats. Intermediate rollout states remain in latent space, and only the final predicted latent is decoded for scoring.

\subsection{Main Results}

Table~\ref{tab:main_results} summarizes quantitative comparisons across four visual‑navigation benchmarks. Our SkipWorld achieves competitive overall ATE and RPE performance against prior world‑model‑based navigation methods. Note that NWM (original) is fully trained for 100K steps from the original work, whereas our NWM (pretrained) baseline only receives an additional 50K‑step fine‑tune, which partially accounts for its performance gap versus NWM (original). Even under this limited training budget, SkipWorld obtains strong average‑tracking metrics and yields clear improvements over our retrained NWM (pretrained) baseline.

\begin{table}[t]
\centering
\caption{Component ablation averaged over four datasets. Time is seconds per query. Lower is better. The best result is shown in bold and the second best is underlined.}
\label{tab:component_ablation}
\small
\begin{tabular}{lccc ccc}
\toprule
Variant & AutoRoll & SkipDistill & Refine & ATE & RPE & Time \\
\midrule
NWM
& -- & -- & --
& 3.33 & 0.73 & 2220.5 \\
Latent AutoRoll
& \checkmark & -- & --
& 3.26 & \underline{0.72} & 3807.9 \\
AutoRoll + SkipDistill
& \checkmark & \checkmark & --
& 3.26 & 0.73 & \textbf{465.3} \\
AutoRoll + Refine
& \checkmark & -- & \checkmark
& \underline{3.20} & \textbf{0.71} & 991.7 \\
\textbf{SkipWorld}
& \checkmark & \checkmark & \checkmark
& \textbf{3.19} & \textbf{0.71} & \underline{834.5} \\
\bottomrule
\end{tabular}
\end{table}

\begin{table*}[t]
\centering
\caption{Ablation on skip stride with candidate budget fixed to $P=20$. Time is seconds per query. Lower is better. The best result is shown in bold and the second best is underlined.}
\label{tab:stride_ablation}
\small
\begin{tabular}{lccccccccccc}
\toprule
\multirow{2}{*}{Stride}
& \multicolumn{2}{c}{RECON}
& \multicolumn{2}{c}{HuRoN}
& \multicolumn{2}{c}{TartanDrive}
& \multicolumn{2}{c}{SCAND}
& \multicolumn{2}{c}{Avg.}
& \multirow{2}{*}{Time} \\
\cmidrule(lr){2-3}
\cmidrule(lr){4-5}
\cmidrule(lr){6-7}
\cmidrule(lr){8-9}
\cmidrule(lr){10-11}
& ATE & RPE & ATE & RPE & ATE & RPE & ATE & RPE & ATE & RPE & \\
\midrule
2 
& 1.55 & 0.39 
& 4.97 & 1.06 
& 5.90 & 1.23 
& 1.06 & \underline{0.29} 
& 3.37 & 0.74 
& 1257.7 \\
4 
& \underline{1.30} & \underline{0.36}
& \underline{4.94} & \underline{1.05}
& \textbf{5.72} & \textbf{1.19}
& \underline{1.02} & \textbf{0.28}
& \underline{3.24} & \underline{0.72}
& \underline{973.7} \\
8 
& \textbf{1.19} & \textbf{0.34}
& \textbf{4.86} & \textbf{1.04}
& \underline{5.74} & \underline{1.20}
& \textbf{0.98} & \textbf{0.28}
& \textbf{3.19} & \textbf{0.71}
& \textbf{834.5} \\
\bottomrule
\end{tabular}
\end{table*}

\begin{table*}[t]
\centering
\caption{Ablation on coarse candidate budget with skip stride fixed to $s=8$. Time is seconds per query. Lower is better. The best result is shown in bold and the second best is underlined.}
\label{tab:budget_ablation}
\small
\begin{tabular}{lccccccccccc}
\toprule
\multirow{2}{*}{Budget}
& \multicolumn{2}{c}{RECON}
& \multicolumn{2}{c}{HuRoN}
& \multicolumn{2}{c}{TartanDrive}
& \multicolumn{2}{c}{SCAND}
& \multicolumn{2}{c}{Avg.}
& \multirow{2}{*}{Time} \\
\cmidrule(lr){2-3}
\cmidrule(lr){4-5}
\cmidrule(lr){6-7}
\cmidrule(lr){8-9}
\cmidrule(lr){10-11}
& ATE & RPE & ATE & RPE & ATE & RPE & ATE & RPE & ATE & RPE & \\
\midrule
$P=5$
& 1.32 & 0.36
& 4.92 & 1.06
& \textbf{5.67} & \underline{1.19}
& 1.07 & 0.29
& 3.25 & 0.73
& \textbf{460.5} \\

$P=10$
& 1.33 & 0.36
& 4.89 & \underline{1.05}
& \textbf{5.67} & \textbf{1.18}
& \textbf{0.96} & \underline{0.28}
& \underline{3.21} & \underline{0.72}
& \underline{610.1} \\

$P=20$
& \underline{1.19} & \underline{0.34}
& \textbf{4.86} & \textbf{1.04}
& \underline{5.74} & 1.20
& \underline{0.98} & \underline{0.28}
& \textbf{3.19} & \textbf{0.71}
& 834.5 \\

$P=40$
& \textbf{1.15} & \textbf{0.33}
& \underline{4.88} & \underline{1.05}
& 5.83 & 1.21
& \underline{0.98} & \textbf{0.27}
& \underline{3.21} & \underline{0.72}
& 1436.1 \\
\bottomrule
\end{tabular}
\end{table*}

\subsection{Ablation Studies}

\noindent\textbf{Does each component contribute?} We first ablate the three main components of SkipWorld: Latent AutoRoll, SkipDistill, and Skip-then-Refine Planning. Table~\ref{tab:component_ablation} shows that Latent AutoRoll improves planning accuracy over NWM, but it is expensive when used as a dense autoregressive planner. SkipDistill greatly reduces runtime, although using the skip model alone is less reliable than using refinement. SkipWorld combines these two behaviors: it stays close to the accuracy of AutoRoll-based refinement while reducing the cost of planning.

\noindent\textbf{How large should the skip stride be?} We further ablate the skip stride for coarse candidate screening. As shown in Table~\ref{tab:stride_ablation}, smaller strides yield negligible accuracy improvements at the cost of more rollout calls. Stride‑8 achieves favorable runtime with near‑peak accuracy, verifying the skip model’s effective long‑horizon ranking capability. We adopt stride 8 in our final SkipWorld setup.
\begin{figure}[t]
    \centering
    \includegraphics[width=\linewidth]{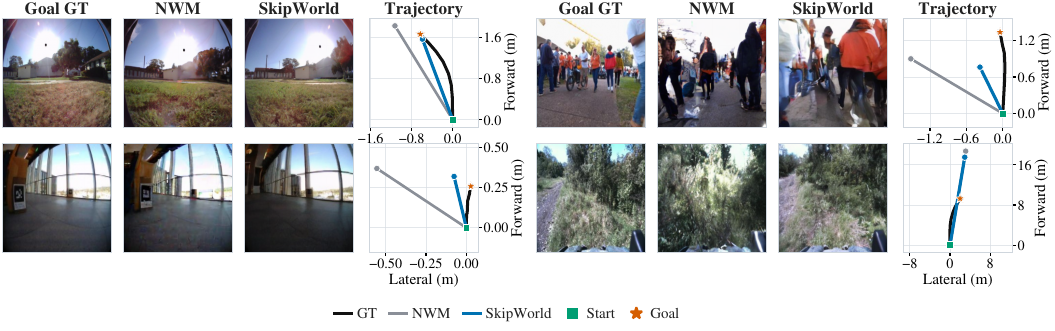}
    \caption{
        Predicted trajectories on the four evaluation datasets. SkipWorld produces trajectories that better follow the ground-truth path across diverse indoor, outdoor, and driving scenes.
    }
    \label{fig:scene_trajectory_cases}
\end{figure}

\noindent\textbf{How many candidates should be retained?} Finally, we vary the number of candidates retained after coarse screening. Table~\ref{tab:budget_ablation} shows the expected trade-off: a small budget is fast but can remove useful candidates too early, while a larger budget gives the refinement stage more choices at a higher cost. The improvement from increasing the budget is not uniform across datasets, which indicates that candidate filtering quality depends on the environment and trajectory distribution. In practice, a moderate budget offers a reasonable balance between preserving good candidates and avoiding excessive teacher evaluation.

\subsection{Qualitative Results}

Figure~\ref{fig:scene_trajectory_cases} visualizes representative predicted trajectories on the four evaluation datasets. Compared with the original NWM planner, SkipWorld tends to produce trajectories that better follow the ground-truth future motion, especially when the original planner selects candidates with noticeable lateral drift. These examples provide qualitative evidence that latent skip scoring does not merely accelerate planning, but can also guide the planner toward more plausible action candidates before final refinement.

\section{Conclusion}

We presented SkipWorld, a latent skip planning framework for efficient visual navigation. SkipWorld first trains a latent autoregressive world model to reduce the mismatch between training and closed-loop inference, then distills it into a skip student that directly predicts long-stride future latent states. During planning, the skip student efficiently filters candidate trajectories, while the autoregressive teacher refines only the selected candidates. Experiments on four navigation datasets show that this design improves the speed--accuracy trade-off over dense world-model planning. These results suggest that latent skip prediction is a practical way to make visual world-model planning more efficient without sacrificing trajectory quality.

\subsection*{Acknowledgments}

This work was supported in part by the InnoHK initiative of the Innovation and Technology Com-mission of the Hong Kong Special Administrative Region Government via the Hong Kong Centre for Logistics Robotics. This work was also supported in part by project MMT-p8-26 of the Shun Hing Institute of Advanced Engineering, The Chinese University of Hong Kong.

\subsection*{AI use statement}

During the preparation of this paper, we used generative AI tools for grammar checking and formatting assistance. All technical contributions, including problem formulation, methodology design, experimental implementation, result analysis, and conclusions, were developed and verified by the authors. The authors take full responsibility for the final content of this work.

\subsection*{Ethics Statement}

This work does not involve human subjects, personally identifiable information, or sensitive data collection. The datasets used in our experiments are publicly available benchmarks for visual navigation research. We do not identify any additional ethical concerns related to the methodology or applications presented in this work.

\subsection*{Reproducibility Statement}

We provide detailed descriptions of our methodology, model architecture, training procedures, and experimental settings in the main paper and supplementary materials. We report the datasets, evaluation protocols, implementation details, and hyperparameters necessary to reproduce our experiments. We will release the source code and configurations to facilitate further research and verification of our results.



\bibliography{iclr2027_conference}
\bibliographystyle{iclr2027_conference}

\appendix
\section{Appendix}

\subsection{Reproducibility and Evaluation Details}

We provide additional implementation and evaluation details for
reproducibility. Unless otherwise stated, all planning experiments use
the same candidate optimization setting summarized in
Table~\ref{tab:appendix_hparams}. The planner samples $N=120$ candidate
action trajectories, retains the top $K=5$ candidates, and performs one
CEM update step. Dense autoregressive scoring is repeated $R=3$ times
to reduce stochastic variance. The default SkipWorld configuration uses
a coarse stride $s_c=8$ and retains $P=20$ candidates for dense
refinement.

\begin{table}[h]
\centering
\caption{Default planning hyperparameters used in our experiments.}
\label{tab:appendix_hparams}
\small
\begin{tabular}{lc}
\toprule
Hyperparameter & Value \\
\midrule
Candidate samples $N$ & 120 \\
Elite candidates $K$ & 5 \\
CEM update steps & 1 \\
Batch size & 1 \\
Number of workers & 12 \\
Dense repeat evaluations $R$ & 3 \\
Coarse repeat evaluations & 1 \\
Fine rollout stride & 1 \\
Default coarse stride $s_c$ & 8 \\
Default preselection size $P$ & 20 \\
Prediction horizon $H$ & 8 frames \\
Evaluation frame rate & 4 FPS \\
\bottomrule
\end{tabular}
\end{table}

All methods are evaluated on the same test split for each dataset. The
default prediction horizon is 8 future frames, corresponding to 2
seconds at 4 FPS. Runtime is measured as the average wall-clock planning
time per query and includes candidate rollout, final latent decoding,
perceptual scoring, and evaluation overhead under the same implementation
setting.

\subsection{Full Component Ablation Results}

The main paper reports the averaged component ablation results. Here,
Table~\ref{tab:appendix_full_component} provides the full dataset-wise
results.

Latent AutoRoll improves average trajectory accuracy over NWM by reducing
the mismatch between training and closed-loop rollout, although dense
autoregressive planning remains computationally expensive. Skip-only
planning substantially reduces runtime by replacing dense rollout with
long-stride prediction during candidate evaluation. Adding dense
refinement improves robustness by reranking the shortlisted candidates
with the stronger autoregressive teacher. Overall, SkipWorld achieves
the best runtime among the refined variants while maintaining comparable
trajectory accuracy.

\begin{table*}[t]
\centering
\caption{Full dataset-wise component ablation. Time is average planning time per query. Lower is better. The best result is shown in bold and the second best is underlined.}
\label{tab:appendix_full_component}
\small
\setlength{\tabcolsep}{4pt}
\begin{tabular}{lccccccccccc}
\toprule
\multirow{2}{*}{Method}
& \multicolumn{2}{c}{RECON}
& \multicolumn{2}{c}{HuRoN}
& \multicolumn{2}{c}{TartanDrive}
& \multicolumn{2}{c}{SCAND}
& \multicolumn{2}{c}{Avg.}
& \multirow{2}{*}{Time} \\
\cmidrule(lr){2-3}
\cmidrule(lr){4-5}
\cmidrule(lr){6-7}
\cmidrule(lr){8-9}
\cmidrule(lr){10-11}
& ATE & RPE & ATE & RPE & ATE & RPE & ATE & RPE & ATE & RPE & \\
\midrule
NWM
& 1.381 & 0.351
& 5.076 & 1.083
& 5.791 & 1.209
& 1.081 & 0.290
& 3.332 & 0.733
& 2220.5 \\

Latent AutoRoll
& \underline{1.245} & \textbf{0.336}
& 4.909 & 1.048
& 5.846 & 1.216
& 1.020 & 0.277
& 3.255 & 0.720
& 3807.9 \\

AutoRoll + SkipDistill
& 1.487 & 0.380
& \underline{4.846} & 1.043
& \textbf{5.638} & \textbf{1.178}
& 1.067 & 0.297
& 3.260 & 0.725
& \textbf{465.3} \\

AutoRoll + Refine
& \textbf{1.191} & \textbf{0.336}
& \textbf{4.839} & \textbf{1.038}
& 5.798 & 1.208
& \underline{0.987} & \textbf{0.271}
& \underline{3.204} & \textbf{0.713}
& 991.7 \\

SkipWorld
& \textbf{1.191} & \underline{0.344}
& 4.859 & \underline{1.039}
& \underline{5.736} & \underline{1.196}
& \textbf{0.980} & \underline{0.277}
& \textbf{3.192} & \underline{0.714}
& \underline{834.5} \\
\bottomrule
\end{tabular}
\end{table*}

\subsection{Horizon-wise Error Analysis}

In addition to the final 2-second planning metric reported in the main
paper, we evaluate trajectory accuracy at intermediate prediction
horizons. Specifically, each saved trajectory prediction is truncated at
the corresponding horizon, after which ATE and RPE are recomputed.
Table~\ref{tab:horizon_error} summarizes the resulting horizon-wise
errors. No additional planning or model inference is performed for this
analysis; all results are derived from the same saved predictions used
in the main experiments.

\begin{table*}[t]
\centering
\caption{Horizon-wise error analysis averaged over four datasets. Lower is better. The best result is shown in bold and the second best is underlined.}
\label{tab:horizon_error}
\setlength{\tabcolsep}{4.5pt}
\small
\renewcommand{\arraystretch}{1.12}
\begin{tabular}{lcccccccc}
\toprule
\multirow{2}{*}{Method}
& \multicolumn{2}{c}{0.5s}
& \multicolumn{2}{c}{1.0s}
& \multicolumn{2}{c}{1.5s}
& \multicolumn{2}{c}{2.0s} \\
\cmidrule(lr){2-3}
\cmidrule(lr){4-5}
\cmidrule(lr){6-7}
\cmidrule(lr){8-9}
& ATE & RPE & ATE & RPE & ATE & RPE & ATE & RPE \\
\midrule
NWM
& 1.134 & 0.708
& 1.884 & 0.706
& 2.607 & 0.716
& 3.332 & 0.733 \\

Latent AutoRoll
& 1.119 & 0.694
& 1.854 & 0.693
& 2.551 & 0.700
& 3.255 & 0.720 \\

AutoRoll+SkipDistill
& \underline{1.092} & 0.682
& 1.821 & 0.687
& 2.532 & 0.701
& 3.260 & 0.725 \\

AutoRoll+Refine
& 1.098 & \underline{0.681}
& \underline{1.816} & \underline{0.682}
& \underline{2.505} & \underline{0.693}
& \underline{3.204} & \textbf{0.713} \\

SkipWorld
& \textbf{1.084} & \textbf{0.673}
& \textbf{1.797} & \textbf{0.677}
& \textbf{2.487} & \textbf{0.691}
& \textbf{3.192} & \underline{0.714} \\

\bottomrule
\end{tabular}
\end{table*}

\subsection{Loss Ablation for SkipDistill}
\label{app:loss_ablation}

We further study the training objective of the skip student in
Table~\ref{tab:loss_ablation}. Given a latent context
$C_t=\{z_{t-K+1},\ldots,z_t\}$ and a skip stride $s$, the student
predicts the future endpoint latent at $t+s$ conditioned on the
accumulated action
$\bar{a}_{t,s}=\sum_{i=1}^{s}a_{t+i}$ and the relative temporal
condition $s\tau$.

\paragraph{Direct endpoint supervision.}
A straightforward choice is to directly supervise the skip student
using the ground-truth future latent $z_{t+s}$. The corresponding
diffusion objective is

\begin{equation}
\mathcal{L}_{\mathrm{direct}}
=
\mathcal{L}_{\mathrm{diff}}
\left(
F_{\phi},
z_{t+s},
C_t,
\bar{a}_{t,s},
s\tau
\right),
\end{equation}

where $F_{\phi}$ denotes the skip student and
$\mathcal{L}_{\mathrm{diff}}$ is the diffusion denoising loss.

\paragraph{Teacher endpoint distillation.}
Alternatively, we supervise the student using the endpoint produced by
the frozen Latent AutoRoll teacher. Starting from the same context
$C_t$, the teacher performs $s$ dense one-step transitions to obtain
$\hat{z}^{T}_{t+s}$. The student prediction
$\hat{z}^{S}_{t+s}$ is then aligned with this teacher endpoint using

\begin{equation}
\mathcal{L}_{\mathrm{distill}}
=
\mathrm{SmoothL1}
\left(
\hat{z}^{S}_{t+s},
\mathrm{sg}\left(\hat{z}^{T}_{t+s}\right)
\right),
\end{equation}

where $\mathrm{sg}(\cdot)$ denotes stop-gradient.

\paragraph{Ablation setting.}
We compare three training objectives:
$\mathcal{L}_{\mathrm{direct}}$,
$\mathcal{L}_{\mathrm{direct}}+\mathcal{L}_{\mathrm{distill}}$, and
$\mathcal{L}_{\mathrm{distill}}$.
All variants are evaluated using the same Skip-then-Refine planner with
skip stride $s=8$ and preselection budget $P=20$, so that the comparison
isolates the effect of the skip-student training objective.

\paragraph{Results.}
As shown in Table~\ref{tab:loss_ablation}, distillation-only achieves
the best average ATE while remaining competitive in RPE.
In contrast, direct endpoint supervision performs worse on average,
indicating that the endpoint generated by the autoregressive teacher
provides a more effective long-stride training target than the
ground-truth endpoint alone. Combining the two objectives does not
provide a consistent improvement over distillation-only.
We therefore use $\mathcal{L}_{\mathrm{distill}}$ as the default
SkipDistill objective in all main experiments.

\begin{table*}[t]
\centering
\caption{Loss ablation of SkipDistill under the same Skip-then-Refine planning setting. Time is average planning time per query. Lower is better. The best result is shown in bold and the second best is underlined.}
\label{tab:loss_ablation}
\small
\setlength{\tabcolsep}{3.0pt}
\begin{tabular}{lccccccccccc}
\toprule
\multirow{2}{*}{Training Loss} &
\multicolumn{2}{c}{RECON} &
\multicolumn{2}{c}{HuRoN} &
\multicolumn{2}{c}{TartanDrive} &
\multicolumn{2}{c}{SCAND} &
\multicolumn{2}{c}{Avg.} &
\multirow{2}{*}{Time} \\
\cmidrule(lr){2-3}
\cmidrule(lr){4-5}
\cmidrule(lr){6-7}
\cmidrule(lr){8-9}
\cmidrule(lr){10-11}
& ATE & RPE & ATE & RPE & ATE & RPE & ATE & RPE & ATE & RPE & \\
\midrule
$\mathcal{L}_{\mathrm{direct}}$
& 1.235 & 0.345
& 4.945 & 1.056
& 5.807 & 1.208
& \underline{0.996} & \textbf{0.275}
& 3.246 & \underline{0.721}
& \textbf{824.5} \\

$\mathcal{L}_{\mathrm{direct}}+\mathcal{L}_{\mathrm{distill}}$
& \textbf{1.168} & \textbf{0.331}
& \underline{4.896} & \underline{1.048}
& \underline{5.779} & \underline{1.201}
& 0.999 & \underline{0.276}
& \underline{3.210} & \textbf{0.714}
& 839.9 \\

$\mathcal{L}_{\mathrm{distill}}$
& \underline{1.191} & \underline{0.344}
& \textbf{4.859} & \textbf{1.039}
& \textbf{5.736} & \textbf{1.196}
& \textbf{0.980} & 0.277
& \textbf{3.192} & \textbf{0.714}
& \underline{834.5} \\
\bottomrule
\end{tabular}
\end{table*}

\subsection{Latent Alignment Error Visualization}
\label{app:latent_alignment_error}

To visualize the endpoint matching learned by SkipDistill, we compute the latent alignment error between the skip student and the autoregressive teacher. Given the same latent context $C_t$ and action sequence, the teacher autoregressively rolls out $s$ one-step transitions to obtain the endpoint latent $\hat{z}^{T}_{t+s}$, while the skip student directly predicts the endpoint latent $\hat{z}^{S}_{t+s}$ using the accumulated action. For each spatial location $(u,v)$ in the latent feature map, we compute
\begin{equation}
E(u,v)
=
\sqrt{
\frac{1}{C}
\sum_{c=1}^{C}
\left(
\hat{z}^{S}_{t+s,c,u,v}
-
\hat{z}^{T}_{t+s,c,u,v}
\right)^2
}.
\end{equation}
The resulting heatmap shows where the student prediction is well aligned with the teacher rollout in latent space. Lower values indicate closer agreement between the long-stride student transition and the teacher's autoregressive dynamics.

\subsection{Additional Qualitative Results}

\begin{figure*}[t]
    \centering
    \includegraphics[width=\textwidth]{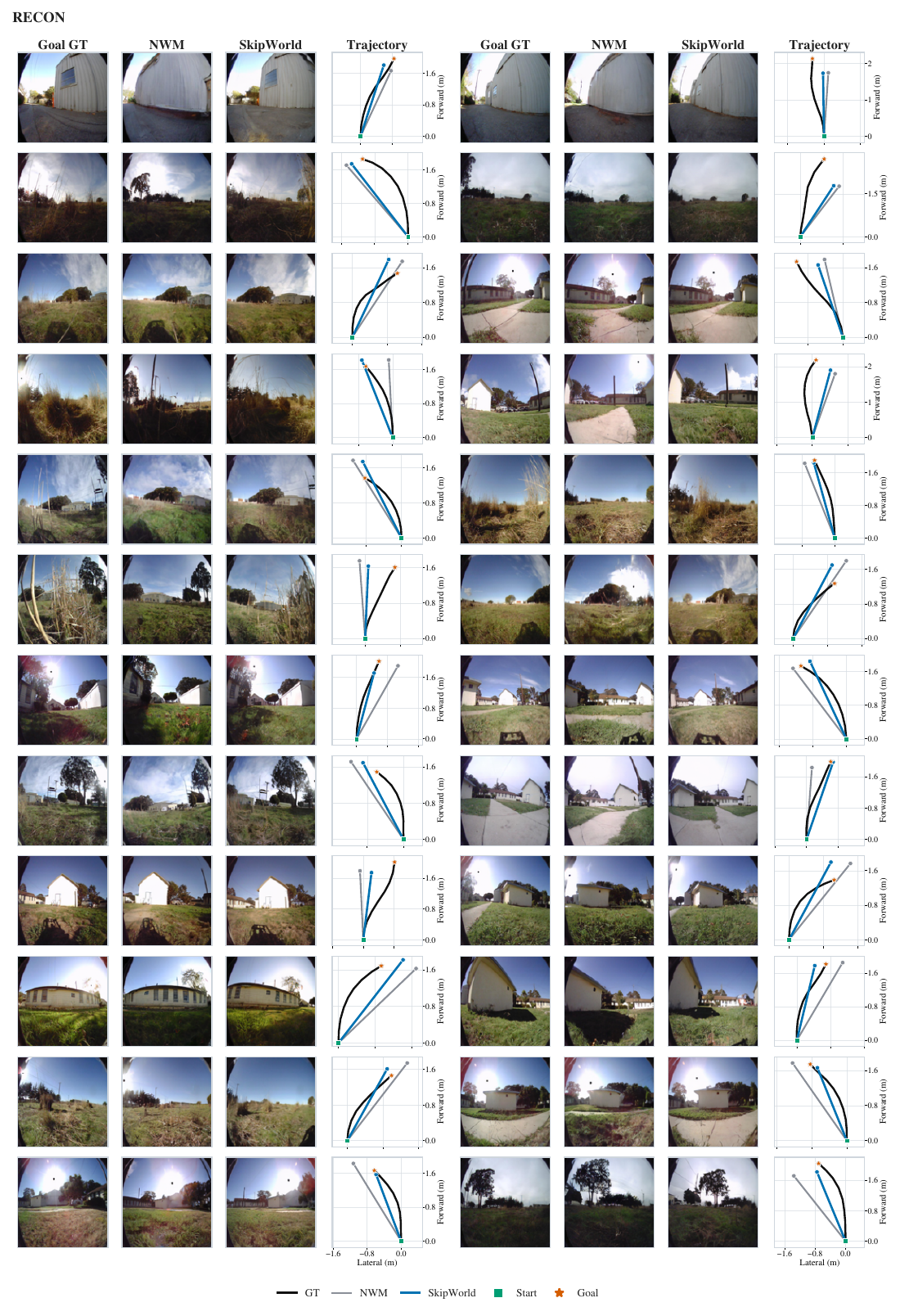}
    \caption{Additional scene-aware qualitative results on RECON.}
    \label{fig:appendix_scene_recon}
\end{figure*}

\begin{figure*}[t]
    \centering
    \includegraphics[width=\textwidth]{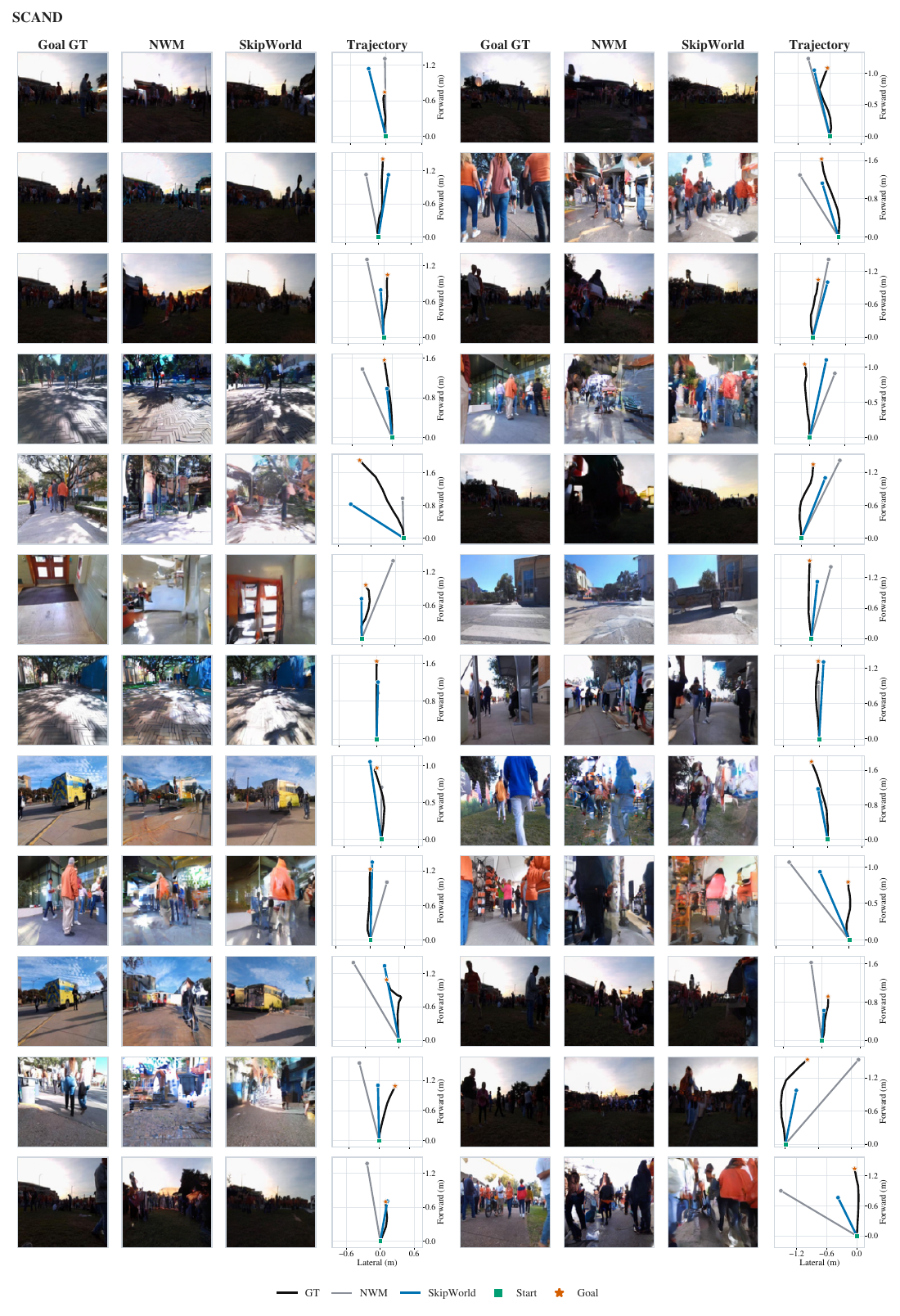}
    \caption{Additional scene-aware qualitative results on SCAND.}
    \label{fig:appendix_scene_scand}
\end{figure*}

\begin{figure*}[t]
    \centering
    \includegraphics[width=\textwidth]{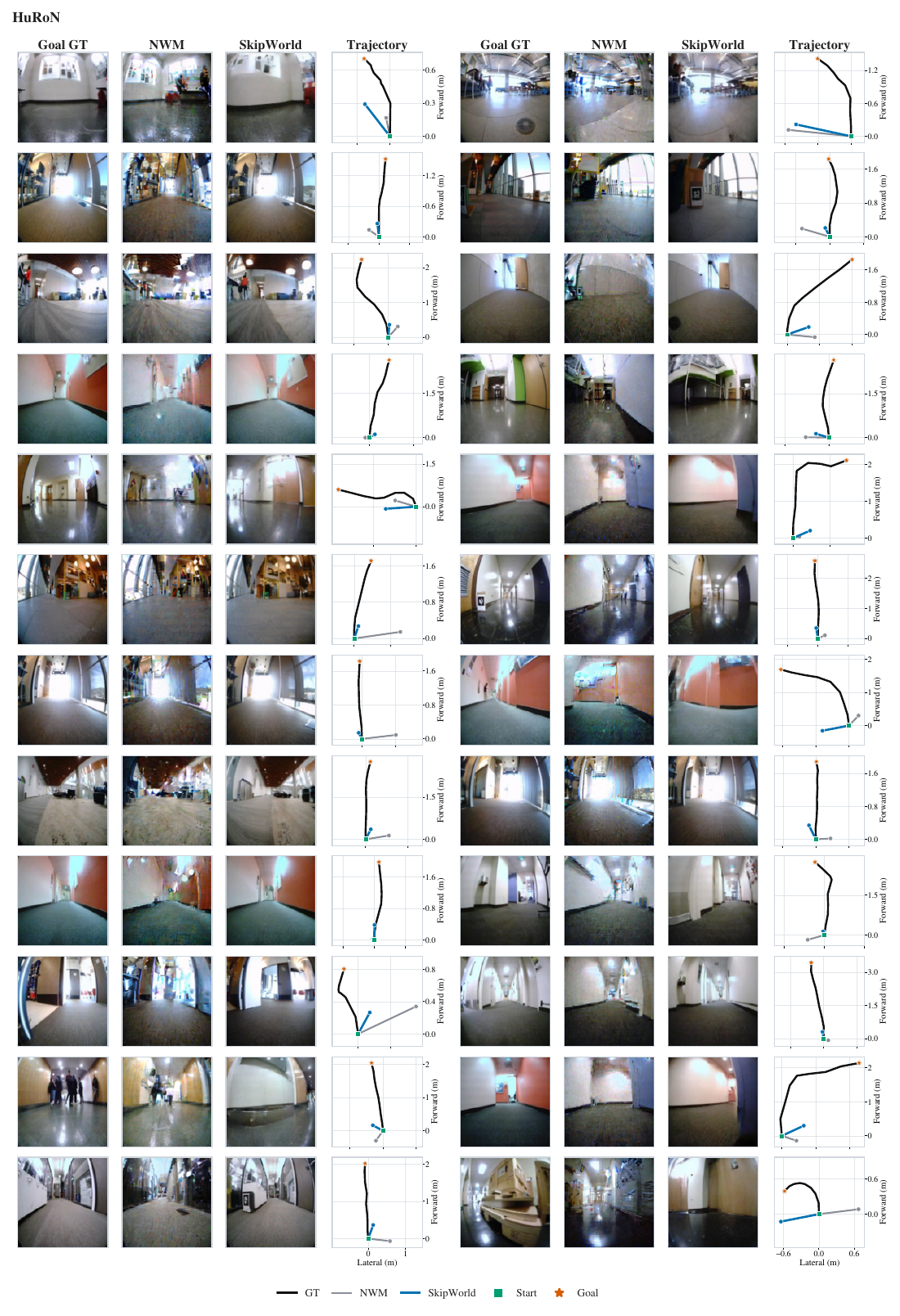}
    \caption{Additional scene-aware qualitative results on HuRoN.}
    \label{fig:appendix_scene_huron}
\end{figure*}

\begin{figure*}[t]
    \centering
    \includegraphics[width=\textwidth]{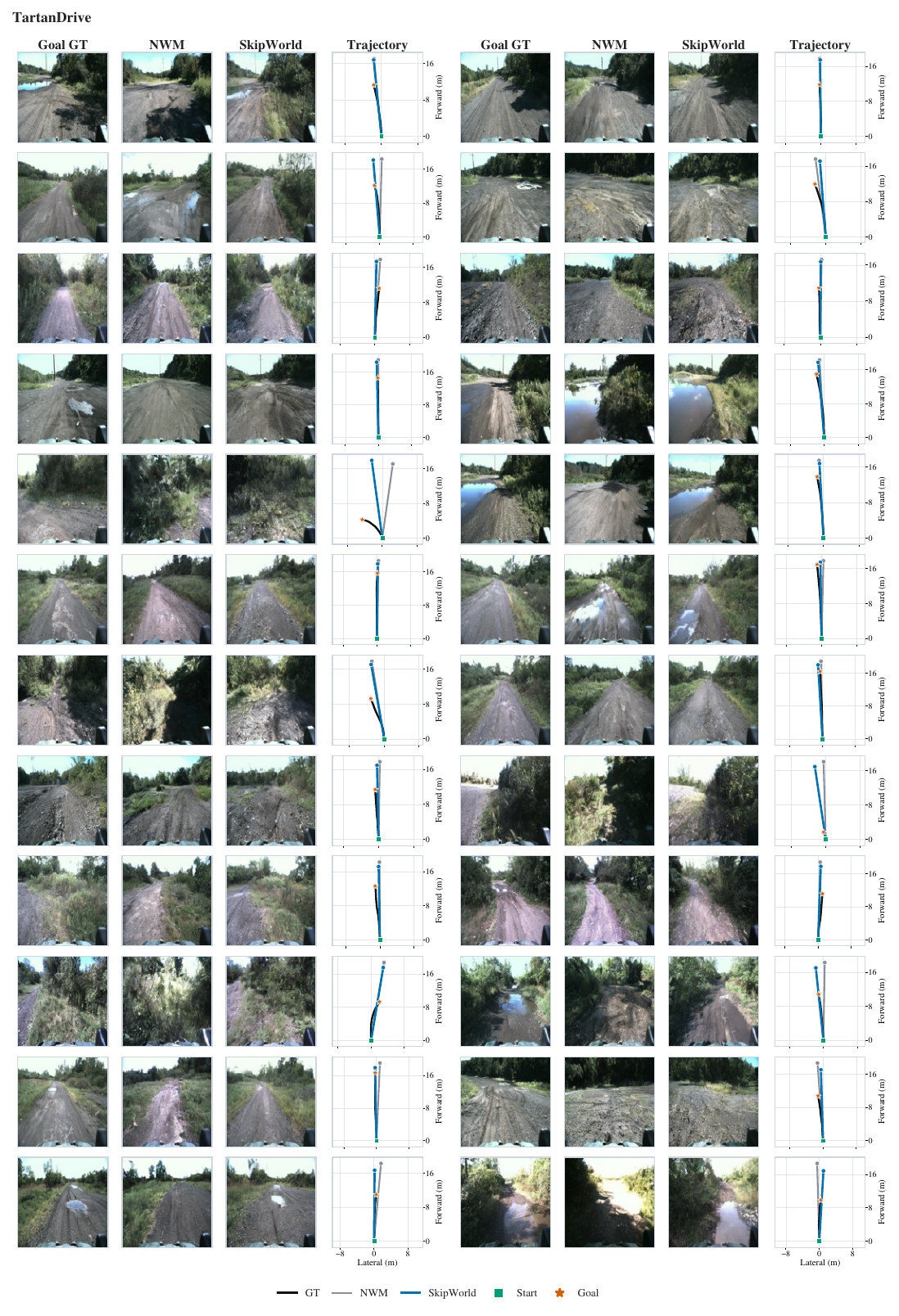}
    \caption{Additional scene-aware qualitative results on TartanDrive.}
    \label{fig:appendix_scene_tartandrive}
\end{figure*}

The main paper presents representative qualitative examples across the
evaluated datasets. Here, we provide additional per-dataset
visualizations for RECON, SCAND, HuRoN, and TartanDrive in
Figures~\ref{fig:appendix_scene_recon},
\ref{fig:appendix_scene_scand},
\ref{fig:appendix_scene_huron}, and
\ref{fig:appendix_scene_tartandrive}, respectively.

Each row contains two planning queries. For each query, we show the goal
image, the predictions produced by NWM and SkipWorld, and the corresponding
planned trajectories. These examples complement the quantitative results by
illustrating how the two planners behave across different visual scenes and
navigation conditions.

\end{document}